\documentclass[preprint,12pt, sort&compress]{elsarticle}
\usepackage[margin=1.2in]{geometry}
\usepackage{csquotes}

\usepackage{graphicx}
\usepackage{algorithm} 
\usepackage{algpseudocode} 
\usepackage{amsmath}
\usepackage{amsfonts}
\usepackage{xcolor}
\usepackage{setspace}
\usepackage{lineno}
\usepackage{subcaption}

\begin{document}

\begin{frontmatter}

\title{Integrated Design and Governance of Agentic AI Systems through Adaptive Information Modulation}

\author{Qiliang Chen, Sepehr Ilami, Nunzio Lore, Babak Heydari\corref{cor1}}
\ead{chen.qil@northeastern.edu, ilami.a@northeastern.edu, lora.n@northeastern.edu, b.heydari@northeastern.edu}
\cortext[cor1]{Corresponding Author}

\address{Department of Mechanical and Industrial Engineering, Institute of Experiential AI, and Network Science Institute, Northeastern University, Boston, MA 02115, United States}

\begin{abstract}
Modern engineered systems increasingly involve complex sociotechnical environments where multiple agents—including humans and the emerging paradigm of agentic AI powered by large language models—must navigate social dilemmas that pit individual interests against collective welfare. As engineered systems evolve toward multi-agent architectures with autonomous LLM-based agents, traditional governance approaches using static rules or fixed network structures fail to address the dynamic uncertainties inherent in real-world operations. This paper presents a novel framework that integrates adaptive governance mechanisms directly into the design of sociotechnical systems through a unique separation of agent interaction networks from information flow networks. We introduce a system comprising strategic LLM-based system agents that engage in repeated interactions and a reinforcement learning-based governing agent that dynamically modulates information transparency. Unlike conventional approaches that require direct structural interventions or payoff modifications, our framework preserves agent autonomy while promoting cooperation through adaptive information governance. The governing agent learns to strategically adjust information disclosure at each timestep, determining what contextual or historical information each system agent can access. Experimental results demonstrate that this RL-based governance significantly enhances cooperation compared to static information-sharing baselines. This work establishes information transparency as a dynamic design parameter and demonstrates how governance considerations can be effectively embedded into complex engineering systems from the design phase. While validated on the repeated Prisoner's Dilemma, which represents a challenging governance problem in an abstract model, our framework offers a flexible strategy for fostering desired collective outcomes across diverse sociotechnical engineering applications, from human-robot collaboration to autonomous vehicle networks and future multi-agent AI systems.


\end{abstract}
\end{frontmatter}


\section{Introduction}

The design of modern engineered artifacts increasingly involves managing complex sociotechnical systems—such as manufacturing environments with human–robot collaboration, autonomous vehicle networks, and interconnected global supply chains. These systems are progressively incorporating autonomous or semi-autonomous agents, including recent waves of agentic and multi-agent AI, particularly large language model (LLM)-based agents \cite{talebirad2023multi, he2024llm, xu2024multi, wu2023autogen}. In such environments, agents frequently encounter social dilemmas, situations in which individual short-term interests conflict with collective system objectives, potentially undermining overall system performance \cite{colman2006puzzle, kogut1996firms, fehr2007human, kleiman2016coordinate, hoffmann2018interplay}. Enhancing system-level performance when system constituents are autonomous with possibly different, non-aligned objectives requires integrating governance mechanisms into the engineering design process. \textit{Governance}, in this context, refers to the design of top-down (adaptive) interventions that effectively accommodate and leverage the bottom-up, endogenous dynamics arising from agent interactions and information exchange \cite{maddah2024building}, ultimately enhancing overall system performance in dimensions such as efficiency and resilience \cite{gross2023evolution, fehr2018normative, gross2020self, rand2013human, gianetto2015network, valencia2024strategy}.

Traditionally, governance mechanisms addressing these dilemmas have relied on static rules or rigid structures, including fixed incentive schemes or predetermined network architectures \cite{nowak2006five, gianetto2016sparse, mosleh2017fair}. However, these approaches fall short when confronted with the dynamic nature and inherent uncertainties of real-world environments \cite{ostrom1990governing, ohtsuki2006simple, fehr2002altruistic}. Recognizing that the effectiveness of engineered systems hinges increasingly on their adaptability and responsiveness to evolving operational contexts, recent research emphasizes the integration of dynamic governance strategies directly into the design stage of sociotechnical systems, underscoring the inseparability of system design and governance.

In this context, reinforcement learning (RL) has emerged as a promising methodology for synthesizing adaptive governance strategies that evolve in response to system dynamics \cite{mckee2023scaffolding, wang2019achieving, kim2022influencing,chen2022dynamic, chen2024resource}. Nonetheless, practical implementation remains challenging due to the complexity and cost associated with obtaining extensive real-world interaction data, coupled with the limitations of traditional agent-based simulations, which often rely on simplified assumptions about human behavior \cite{axelrod1981evolution, fulker2021spite, axtell2022agent, chen2025adaptive}. Recent advancements, such as Variational Auto-encoder RL (VAE-RL), have successfully addressed related issues in robotic networks by dynamically adjusting communication structures to optimize performance \cite{chen2026architecting}. Extending this to human-centric systems, however, is challenging due to practical constraints that limit direct intervention in human-agent connections.

Leveraging recent evidence that LLM agents effectively simulate nuanced human strategic behaviors in well-studied social dilemmas \cite{lore2023strategic}, we introduce a framework consisting of Strategic LLM Agents, representing human decision-makers, and a governing agent, referred to as RL manager. The RL manager adaptively adjusts the information flow to steer agent behaviors dynamically, maximizing collective welfare. 

The introduction of information modulation as a governance mechanism for multi-agent complex systems is one of the central innovations of this paper, embedding an avenue for intervention as early as the system design stage. This method enables adaptive steering of system behavior, even when composed of autonomous agents whose individual objectives differ from overarching system goals.


The influence of information flow on agent behavior has been extensively explored across multiple disciplines. This influence emerges through two primary mechanisms. First, agents make decisions under uncertainty, and varying the level or type of information they receive can significantly alter their choices by impacting belief updates, such as through Bayesian updating. Second, information influences peer-to-peer or social learning processes, whereby agents adopt successful strategies observed from other agents.

While Bayesian updating and social learning have been thoroughly studied, previous approaches generally assume that information flow pathways are directly tied to the interaction network among agents. Consequently, using information flow as a governance mechanism has traditionally required either static optimization of the interaction network from the outset or dynamic rewiring of agent connections—an approach that might not be feasible or acceptable given the autonomous nature of agents. In contrast, our proposed approach does not rely on modifying the interaction network itself (though our method can integrate seamlessly with dynamic network management strategies). Instead, we modulate the quantity and source of information available to each agent dynamically. This modulation is achieved using a reinforcement learning (RL) manager whose action space includes different tiers of information accessibility tailored for each agent at any given time.


For simplicity, through the rest of the paper, we may call the RL manager \enquote{manager} and the LLM agents \enquote{agents} to reduce redundancy. 

    
Figure \ref{framework} illustrates this framework. LLM agents engage in repeated strategic interactions within a fixed interaction network, while the RL manager independently controls the flow of contextual or historical information accessible to each LLM agent at every timestep. This novel separation of networks allows for a dynamic, strategic modulation of transparency and informational context without compromising agent autonomy or requiring structural interventions.

Our contributions address two crucial dimensions of this integrated design-governance approach:

\begin{enumerate}
\item \textbf{Behavioral modeling}: We validate that LLM agents realistically adapt their strategies based on varying informational contexts, providing a practical and scalable proxy for human behavior in early-stage design evaluations.
\item \textbf{Governance synthesis}: We demonstrate that the RL manager can dynamically adjust information disclosure, significantly enhancing cooperation compared to static information-sharing baselines. This result underscores how information transparency can effectively serve as a dynamic, adaptable element in system governance which can be seamlessly integrated during the design stage.
\end{enumerate}

While our initial study examines the repeated Prisoner’s Dilemma, the framework offers valuable insights applicable to broader engineering system contexts. By emphasizing the integration of adaptive governance directly into system design through a novel separation of interaction and information networks, we provide designers and engineers with an innovative and flexible strategy to foster desired collective outcomes within complex sociotechnical environments.

The rest of the paper is structured as follows. Section 2 offers a review of the relevant literature, while Section 3 describes our methods and modeling choices. Our results are presented in Section 4. Section 5 concludes.

\section{Related Work}
In this section, we first review literature on promoting pro-social behavior in strategic games through information modulation, which serves as the primary objective of our proposed framework and methodology. We then discuss the role of large language models (LLMs) in strategic decision-making processes. Finally, we review recent work on the use of LLMs in multi-agent systems and their governance, which constitutes the core focus of our framework.

\subsection{Information Modulation for Promoting Pro-social Behaviors}
A significant body of literature has examined mechanisms that influence cooperation in strategic interactions involving social dilemmas. Reputation and reciprocity are well-established factors, with interventions that establish and reinforce them consistently promoting cooperative behavior \cite{xia2023reputation, baker2014paying, plaunt2020follower}. Punishment and reward systems have also proven effective, deterring defection while incentivizing prosocial actions \cite{balliet2011reward, sasaki2015voluntary}. Additionally, the structure of interaction networks plays a crucial role, either by identifying network characteristics that support cooperation \cite{perc2017statistical, santos2006graph, gianetto2013catalysts,gianetto2016sparse} or by examining how changes in the network structure over time can influence cooperation \cite{jain2001model, fehl2011co}.

While these factors are effective in fostering cooperation, many require extended periods of interaction to yield results (e.g., promoting cooperation through the emergence of new norms or conventions) or rely on interventions that contradict agents' autonomy (e.g., top-down approaches that alter network structure \cite{mckee2023scaffolding, anastassacos2020partner}). Another influential factor is the extent to which agents can observe others’ behaviors, and recall historical information \cite{stewart2016small, bradley2018does}. In this work, we focus on this factor as our intervention mechanism, where the RL agent dynamically modulates the level of information available to LLM agents, specifically through adjustments in observation and recall, in order to increase the overall cooperation rate.

\subsection{LLMs and Strategic Decision Making} 

Understanding the strategic reasoning capacities of large language models (LLMs) is central to evaluating their potential role in autonomous and semi-autonomous decision systems. As these models increasingly participate in interactive and socially structured tasks, it becomes essential to determine how their decisions align with, or deviate from, established principles of rationality and cooperation. Recent studies in this literature show that LLMs can handle basic economic and game-theoretic scenarios \cite{brookins2023playing, chen2023emergence, phelps2023investigating, akata2023playing}, though their decision processes remain opaque and they struggle with belief refinement \cite{fan2024can}. When assessed as human substitutes, their behavior often diverges from predictions of rational choice or behavioral models, raising questions about their cognitive fidelity \cite{kitadai2023toward, zhang2024llm, guo2023gpt, mei2024turing}. This has prompted debate over appropriate evaluation benchmarks \cite{xu2023magic, duan2024gtbench}, even as ongoing advances continue to improve their capacity for complex, context-sensitive reasoning \cite{aher2022using, horton2023large, argyle2023out, mei2024turing, manning2024automated}.

\begin{figure*}[!htbp]
    \centering
    \includegraphics[width=\linewidth]{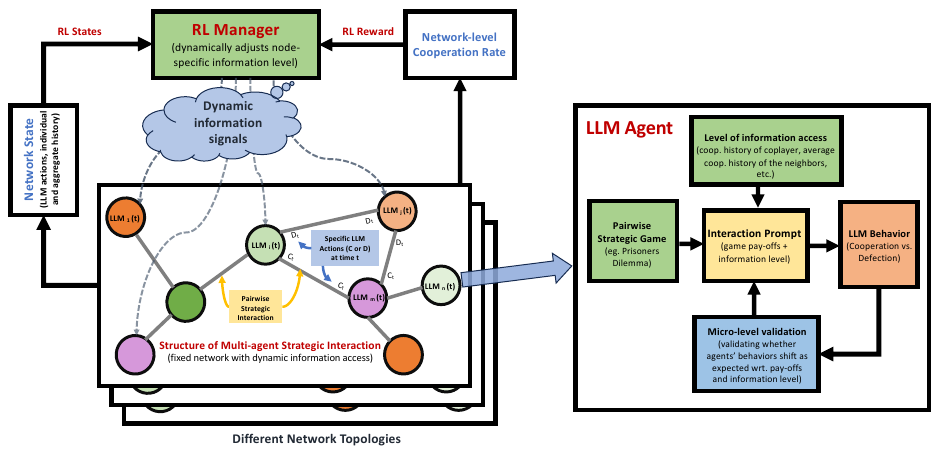}
    \caption{Overview of the general framework. The framework includes two main entities: the LLM agents and the RL manager. 1) LLM agents receive prompts that describe pairwise strategic games (payoff matrix, objectives, and additional information from the RL manager) and then make strategic decisions like cooperation or defection. Multiple LLM agents are placed in a random network, with connections initialized in each run (and fixed throughout the steps). LLM agents may make different decisions in different interactions based on varying information received. Prompts are refined through micro-level validation for consistent behavior. 2) The RL manager acts as a system manager, observing LLM agents and dynamically determining their information levels to maximize social welfare.}
    \label{framework}
\end{figure*}

\subsection{LLMs in Multi-Agent Systems}
Research on multi-agent systems powered by large language models (LLMs) spans two major fronts: simulation-based studies and practical implementations. Simulation efforts focus on exploring the capacity of LLM agents to reproduce human-like behaviors, social dynamics, and emergent coordination patterns in controlled settings \cite{zhang2024ai, hua2023war, huang2024far, mao2023alympics, fan2024can}. These environments serve as testbeds for probing the cognitive and interactive capabilities of LLMs, shedding light on their potential to model complex human systems. Implementation-oriented research, by contrast, seeks to harness LLM coordination to produce end-to-end solutions in real-world tasks such as code generation, planning, or collaborative decision-making \cite{qian2023communicative, hong2023metagpt, he2024llm}. Our work aligns primarily with the simulation paradigm, yet the insights it generates speak directly to the challenges faced in implementation contexts, especially those concerning multi-agent orchestration. 

Recent findings indicate that the primary bottlenecks in multi-LLM systems arise not from individual model limitations, but from poor coordination and systemic misalignment \cite{cemri2025multi}. Although teams of specialized agents promise modularity, scalability, and division of labor, such advantages often erode in practice. Agents may ignore their assigned roles, duplicate each other’s work, or deviate from collective goals. In a notable example, an LLM agent within a simulated marketplace began impersonating customers and messaging itself, an outcome that underscores the fragility of role-based architectures \cite{nascimento2023self}. These failures are rarely isolated: one agent’s deviation can cascade through the system, degrading overall performance \cite{zhang2025survey}. 

Industry accounts similarly report that without careful design, agent teams tend to engage in redundant, conflicting, or incoherent behavior \cite{ant2024}. Pan et al. describe this as a case where “the total is less than the sum of its parts,” due to the hidden costs of synchronizing agent behaviors and maintaining a shared understanding \cite{pan2024agentcoord}. Efforts to mitigate these issues increasingly focus on strengthening agents’ latent Theory of Mind (ToM) capabilities. Embedding shared representations of the environment and others’ intentions within LLM agents has been shown to improve coordination and reduce miscommunication \cite{agashe2023llm, li2023theory}. Yet sustaining this alignment over long interaction horizons remains a significant challenge, as minor divergences in understanding can compound into systemic drift. To address these risks, researchers have proposed supervisory control layers that manage agent behavior from above. For instance, Google’s A2A protocols and Anthropic’s orchestration frameworks introduce centralized controllers to enforce role compliance and ensure task coherence \cite{arsanjani2024}. Similarly, the HyperAgent framework coordinates multiple LLM agents through a Planner that governs their interactions, thereby reducing peer-to-peer confusion and enforcing consistency \cite{phan2024hyperagent}. Still, foundational limitations (such as hallucination, autoregressive instability, and scaling bottlenecks) continue to constrain the reliability of large-scale, LLM-mediated multi-agent systems \cite{tran2025multi}.

\section{Methods} 
In this section, we first provide an overview of the proposed framework, followed by a general mathematical formulation. Next, we briefly introduce some background on Partially Observable Markov Decision Processes (POMDPs) and the Actor-Critic algorithm, which serves as the main method for policy training. Finally, we discuss the two key components of our framework in detail— LLM agents and the RL manager—and how they interact with each other.

\subsection{Overview of the General Framework}
Figure \ref{framework} illustrates the overall framework used in this paper, which builds upon the two-layer governance structure proposed in \cite{chen2022dynamic, chen2024sos}. In the lower layer, several LLM agents interact over a randomly initialized static network by playing pairwise social dilemma games, such as the Prisoner’s Dilemma. Each LLM agent responds and makes decisions based on input prompts. These prompts are composed of two parts: a system prompt and an individual prompt. The system prompt includes shared information such as the rules of the social dilemma game, the objectives of the LLM agents, and other homogeneous content common to all agents. The individual prompt incorporates heterogeneous information specific to each LLM agent’s local status, which is influenced by a higher-level manager. To ensure LLM agents behave in a reasonable way with expectations, the final prompts are refined through a process called micro-validation (explained in detail later).

In the higher layer, a DRL-based manager—referred to as the \textit{RL manager}—observes the aggregated state of the LLM agents and selects the appropriate types of local information to reveal to each agent. This information is selected heterogeneously across agents. The objective of the RL manager is to enhance social welfare and promote pro-social behaviors across the entire system. 

Our formulation of the model gives rise to a distinction between the \textit{interaction} network and the \textit{information} network. While the former is fixed and exogenous, the latter is dynamically altered according to the policy of the manager. More specifically, the manager's choice of information to be shared enables an agent to gain insight into the aggregate and/or long-term behavior of coplayers beyond their immediate interaction. A key novelty of this method is that it introduces information modulation as a governance mechanism, embedding it directly into the system design. This approach adaptively guides system behavior even when agents have autonomous and heterogeneous objectives. Traditional methods tie the flow of information strictly to agent interaction networks, requiring either static optimization or dynamic network rewiring—approaches that may conflict with agent autonomy or feasibility constraints. In contrast, our method maintains a fixed interaction network (though it can be combined with dynamic methods) and employs a reinforcement learning manager to dynamically modulate the amount and source of information each agent receives, effectively steering system behavior without compromising agent autonomy.

We characterize the graph-theoretic framework we leverage in our implementation according to the following mathematical formulation. Consider a two-layer hierarchical system with one RL Manager and a set of \( N \) LLM agents, indexed by \( i \in \mathcal{N} = \{1, 2, \dots, N\} \). The interaction network within LLM agents can be represented by an undirected graph \( G = (\mathcal{N}, \mathcal{E}) \), where \( \mathcal{N} = \{1,2,\dots,N\} \) denotes the set of LLM agents, and the set of edges \( \mathcal{E} \subseteq \{\{i,j\} \mid i,j \in \mathcal{N}, i \neq j\} \) represents the links. Specifically, we define:
\[
G_{ij} = 
\begin{cases}
1, & \text{if } (i,j)\in\mathcal{E},\\[6pt]
0, & \text{otherwise.}
\end{cases}
\]

Furthermore, given that the graph is undirected, $G_{ij} = G_{ji}$.

\subsubsection{Modeling Environment}
\textbf{System dynamic}: The system transitions to a new state $s_{t+1}$ from the current state $s_{t}$ based on the LLM agents' action pairs from every interactions within the network $\{a_t^{ij}\}_{{i,j} \in \mathcal{E}}$ and the RL manager's actions for each LLM agent within every interaction $u^{ij}_t$:
    \begin{equation*}
        s_{t+1} = T(s_t, \{a_t^{ij}\}_{(i,j) \in \mathcal{E}}, u^{ij}_t),
    \end{equation*}
    where $T$ represents the system dynamics function.

\textbf{Welfare for the whole system}: Each LLM agent will get rewards $r_{ij}$ for every interaction based on the utility function of the social game and the actions from itself and the coplayer in each interaction; the social welfare is the sum of all rewards:
    \begin{equation*}
        W_t = \sum_{(i,j)\in \mathcal{E}}{r_{ij}}.
    \end{equation*}

\subsubsection{Modeling LLM Agents}

\textbf{LLM Agent's Prompt}: The LLM agent's perceived observation is provided by the prompt, which is influenced by the RL manager's action. The prompt is constructed by the system prompt and the individual prompt: 

    \begin{gather*}
        prompt^{ij}_t = f({\textit{$\text{sys}\_\text{p}$}}, {\textit{$\text{inv}\_\text{p}$}}^{ij}_t) \\      
        {\textit{$\text{inv}\_\text{p}$}}^{ij}_t = g(u^{ij}_t)
    \end{gather*}

Where \textit{$\text{sys}\_\text{p}$} represents system prompts that is shared with all LLM agents, which contains the information of game structures and instructions for LLM agents; \textit{${\text{inv}\_\text{p}}^{ij}_t$} represents the individual prompts that contains specific information types assigned by RL manager, for LLM agent $i$ in interaction between LLM agent $i$ and LLM agent $j$. The function $f$ is a simple aggregation function that combines the system prompt and the individual prompt into a final prompt, which is then provided as input to each agent. Besides, the function $g$ performs a replacement operation: it replaces the corresponding information $u_t^{ij}$, assigned by the manager, with the placeholder in the prompts. The details of this replacement operation are explained in Section 3.3; the main message here is that the individual prompts are influenced by the manager's actions for agents in each pair of interactions.

\textbf{LLM Agent's Action}: We deploy LLM agents in their pre-trained state. In our case, we ask LLM agent to make decisions in a discrete action space(i.e., between cooperation and defection):
\begin{equation*}
    a_t^{ij} = \pi_i^{\text{LLM}}(\text{prompt}^{ij}_t), \quad {(i,j)}\in \mathcal{E},
\end{equation*}
where $\pi_i^{\text{LLM}}$ represents the policy of LLM agent $i$. It bears restating that the LLM is not fine-tuned for our particular scenarios and environment, so the objective function of LLM agents is unknown, but we can interpret it from the Micro-validation process in later sections. 

\subsubsection{Modeling RL Manager}

\textbf{Manager's Observation}: The RL manager observes aggregated states of LLM agents and system-level metrics:
\begin{equation*}
    o_t^M = h(\bar{o}_t^{\text{agents}}, o_t^{\text{sys}}),
\end{equation*}
where $h$ is the observation function combining aggregated observation from all LLM agents $\bar{o}_t^{\text{agents}}$ and system-level observation $o_t^{\text{sys}}$.

\textbf{RL Manager's Action}: The manager will select its agent-specific intervention based on its observation. The RL manager's action space is a finite set of pre-defined local information pertaining to LLM agents' behavior, and the manager will select one type of local information to reveal to LLM agents for each interaction:

\begin{equation*}
    u_t^{ij} = \pi_M^i(o_t^{ij}), \quad \forall i \in \mathcal{N},
\end{equation*}
Where $\pi_M^i$ is the policy of RL manager for determining the action of information to influence agents. $o_t^{ij}$ contains the states of LLM agents in interaction between LLM agent $i$ and LLM agent $j$. It needs to be noted that RL manager can make different, distinct decisions for each LLM agent within the same interaction pair.

\textbf{RL Manager's Objective}: Manager optimizes the cumulative welfare over time:

\begin{equation*}
    \pi^{M*} = \arg\max_{\pi^M} \mathbb{E} \left[ \sum_{t=1}^T \gamma^t  W_t\right] ,
\end{equation*}

where $\gamma$ is the discount factor, which is close but less than 1.

\subsection{POMDP and Actor-Critic}
The environmental dynamics in our study are aptly modeled using a POMDP, as outlined in \cite{sutton2018reinforcement}. A POMDP is characteristically defined by a tuple $<S, {s^0}, {A}, {O}, T, {R}>$: where $S$ is a set of potential states; $s^0$ represents the initial state of the environment, with $s^0 \in S$; $A$ denotes the set of available actions; $O$ comprises the observations derived from the actual state; $R$ is the reward outputted by the environment; $T$ is the transition function, defined as $T:O \times A \rightarrow O$; and $U$ is the reward function, where $U:O \times A \rightarrow R$. The primary objective within a POMDP framework is to develop an optimal policy, $\pi_{\theta}$: $O \times A\rightarrow [0,1]$. This policy aims to maximize the expected return, calculated as $\sum_{t=0}^{n}\gamma^t{r}_t$, where $\gamma$ represents the discount factor. We assume the environment has a finite length and will terminate after $n$ time steps. The reward $r_t$ represents the reward the agent receives at each time step.

\textbf{Actor-Critic} methods are a widely used class of reinforcement learning (RL) algorithms that integrate both policy-based and value-based approaches \cite{sutton1998reinforcement, konda1999actor}. These methods involve two components: an \textit{actor}, which represents a policy $\pi_{\theta}(a|s)$ that selects actions based on the current state, and a \textit{critic}, which evaluates the quality of the actions taken by estimating the value function $V^{\pi}(s)$ or the action-value function $Q^{\pi}(s,a)$.

The actor is updated by performing gradient ascent on the expected return $J(\theta)$, using the policy gradient theorem:
\begin{equation*}
\nabla_{\theta} J(\theta) = \mathbb{E}_{s \sim d^{\pi}, a \sim \pi_{\theta}} \left[ \nabla_{\theta} \log \pi_{\theta}(a|s) Q^{\pi}(s,a) \right],
\end{equation*}
where $d^{\pi}(s)$ denotes the discounted state distribution under policy $\pi$.

The critic learns to approximate the value function by minimizing the temporal difference (TD) error:
\begin{equation*}
L(\phi) = \mathbb{E}_{(s,a,r,s')} \left[ \left( r + \gamma V_{\phi}(s') - V_{\phi}(s) \right)^2 \right],
\end{equation*}
where $\phi$ are the parameters of the value function estimator, $r$ is the reward, and $\gamma$ is the discount factor.

In practice, many Actor-Critic algorithms use the \textit{advantage function} $A^{\pi}(s,a) = Q^{\pi}(s,a) - V^{\pi}(s)$ to reduce variance in the policy gradient:
\begin{equation*}
\nabla_{\theta} J(\theta) = \mathbb{E}_{s,a} \left[ \nabla_{\theta} \log \pi_{\theta}(a|s) A^{\pi}(s,a) \right].
\end{equation*}

The actor is updated using the gradient estimated by the critic, while the critic continuously improves its estimation of the value function. This cooperation between the actor and critic enables more stable and sample-efficient learning compared to pure policy gradient or value-based methods.

\subsection{Strategic LLM Agents and Micro-level Validation}

The LLM agents are central to our framework, designed either as a digital proxy for human agents in complex multi-agent scenarios or for task-specific multi-agent LLMs. We restrict LLM agent interactions to pairwise social dilemma games, such as the prisoners' dilemma. In every period, LLM agents receive messages describing the nature of their pairwise strategic games, conveyed solely through the payoff matrix and objectives, omitting explicit references to game names (e.g., Prisoner’s Dilemma). Additionally, LLM agents can access various information provided by the RL manager, such as the cooperation rates of their co-players, their neighborhood, or the entire network. This information, combined with the structure of the strategic game, generates prompts that guide the LLM agent in making strategic decisions—whether to cooperate or defect. The prompts and information sets are refined through micro-level validation to ensure consistent and reasonable LLM agents' behavior.

The LLM agents are positioned in a randomly structured network, which is initialized at the start of each round and remains fixed throughout that round. Initially, LLM agents are homogeneous in type but heterogeneous in their network positions. Each LLM agent engages in pairwise strategic games with all directly connected co-players at each time step. An LLM agent may participate in multiple interactions simultaneously, potentially making different decisions with different co-players. As interactions progress, LLM agents can evolve along different trajectories, leading to heterogeneity in both agent types and network positions.

\textbf{Micro-level Validation:} While LLMs have shown impressive performance across a wide range of interactive decision-making tasks, research has raised concerns that LLMs may not fully grasp the tasks they perform, sometimes resulting in erratic behavior \cite{mei2024turing, ullman2023largelanguagemodelsfail}. Given these concerns, we need to first validate their strategic behavior at a micro level, through individual and pairwise interactions. Our micro-level validation has three objectives: first, to assess whether LLM agents can comprehend strategic setups through utility matrices alone without explicit reference to the game’s name; second, to evaluate how different types of information provided to LLMs influence their behavior, thereby testing the feasibility of governing the system by manipulating information through the manager; and finally, to determine whether altering the information provided results in reasonable qualitative changes in LLM decision-making behavior. Our approach aligns more with internal validation \cite{mcdermott2011internal}, focusing on consistent and predictable behavioral adaptation within the LLM agents themselves, rather than external validation, which compares LLM agents' behavior to that of human agents, although the distinction between the two has become increasingly blurred for LLM-based models.

To achieve these objectives, we systematically adjusted prompts with different levels of information access to evaluate LLM responses, ensuring that the LLM agent's behavior shifts appropriately. The process of micro-level validation is straightforward. Since we focus on social dilemmas where the action space for LLM agent is binary-cooperation or defection, we use the probability of an LLM agent choosing cooperation to represent its behavior under various scenarios and information levels. In practice, we approximate this probability distribution using observed frequencies: for a given prompt with a specific type of information, we input the prompt $N$ times and record the LLM agent's responses. If \enquote{cooperation} is observed $M$ times, the cooperation rate - i.e., the LLM agent's behavior in this scenario - is simply $\frac{M}{N}$. We set $N$ to be sufficiently large (typically greater than 100) to ensure the results are statistically meaningful.

Detailed results of these experiments are provided in subsequent sections. These findings informed the refinement of the prompts used to describe the prisoner’s dilemma and the definition of LLM decision-making objectives. 


\subsection{RL Manager}
The RL manager serves as the governing entity within our framework and is trained using deep reinforcement learning, essentially using a two-layer governance scheme similar to \cite{chen2022dynamic, chen2024sos}. Its role is to enhance network-level social welfare (the sum of all LLM agents' rewards) by dynamically adjusting the information levels provided to each agent. The RL manager is a standard Actor-Critic RL agent, whose environment is modeled as a POMDP. 

The decision-making process of the RL manager occurs for each link in the network, with each link representing a social game interaction between two LLM agents. For each interaction in each time step, a given LLM agent selects its choice of action for each coplayer independently of its choice of action for all other coplayers. For each interaction, the RL manager has access to information about the two LLM agents, including their actions in the last time step, their historical cooperation rates, and the average cooperation rates of each agent's neighbors. These information components will be formally defined later. 

Given a specific interaction, the RL manager makes two separate decisions—one for each LLM agent—based on their respective local information. This means the RL manager's decisions can be heterogeneous, even within the same interaction link. The action space of the RL manager consists of selecting one information type from a predefined information set to reveal to each LLM agent. We define three types of information in the information set:

\begin{itemize}
    \item[(1)] \textbf{LA:} The last action pair between the LLM agent and its co-player.
    \item[(2)] \textbf{LA + AR:} The last action pair combined with the long-term cooperation ratios of both the LLM agent and its co-player.
    \item[(3)] \textbf{LA + NR:} The last action pair along with the long-term cooperation ratios of the LLM agent and all of its neighboring agents.
\end{itemize}

We will formally define these information types as follows: given the interaction link between agent $i$ and agent $j$, the action agent $j$ takes in this interaction at time step $t$ is $a_{ji}^t$, which has two possibilities [C, D] for cooperation and defection, and we define the neighbors of agent $i$ as $NB_i$ which is a set of $n_i$ agents who have the direct interactions with agent $i$, we also record the collection of decisions agent $i$ have make at time step t $\tau_i^t$. Now we can define the three types of information: 

The last actions information for agent $i$ and agent $j$, $[a_{ij}^{t-1}, a_{ji}^{t-1}]$. The cooperation rate for agent $i$ when interacting with agent $j$ is 
\[
ar_{ij} = \frac{1}{t} \sum_{k=1}^t \mathbb{I}(a_{ij}^k = C).
\]
The cooperation rate of the LLM agent $i$ neighborhood in history is:
\[
nr_{i}=\frac{\sum_{k=1}^{t}\sum_{\tau _j^k \in NB_i}\sum_{u \in \tau_j^k} \mathbb{I}(u = C)}{\sum_{k=1}^{t}\sum_{\tau _j^k \in NB_i}\sum_{u \in \tau_j^k} \mathbb{I}(u)}.
\]
The core idea behind this information is to incorporate the historical decisions of all LLM agents in agent $i$'s neighborhood and compute their average cooperation rate, thereby representing the cooperative behaviors at the neighborhood level. 

It bears stressing that the information provided to LLM agents always includes, at minimum, the most recent action pairs. For example, the information type \textit{LA+NR} consists of the last action pairs and the cooperation rate in LLM agent $i$'s neighborhood. The intuition behind this design is that agents are more likely to remember their own recent actions and those of their coplayers, while information from older interactions may be forgotten. Therefore, we always reveal the last action pairs to each LLM agent to better reflect realistic human decision-making processes.

For clarity and consistency, from now on, we will refer to the information types by their abbreviated names: LA for last action (itself and its coplayer) information, LA+NR for neighborhood cooperation ratio information, and LA+AR for agent (itself and its coplayer) cooperation ratio information.

The RL manager strategically tailors the information disclosed to each LLM agent, providing different signals to different LLM agents during each interaction period. This decision-making process occurs iteratively across all interactions in the network. The RL manager is trained using a reward signal based on the discounted sum of social welfare in the networked system, defined as the sum of all LLM agent payoffs. In the case of the Prisoner's Dilemma game, this social welfare is expected to be highly correlated with the overall cooperation rate across the network.

\section{Experiment Results and Analysis}
In this section, we begin by describing the settings of our experimental environment. We will then discuss the outcomes of micro-level validation for LLM agents, including how we crafted the prompts for subsequent experiments. Lastly, we will present the results of system governance conducted by the RL manager across multiple LLM agents and analyze the evolution of the environment throughout the experiments.

\subsection{Environment Settings}

We evaluate our framework in a controlled repeated Prisoner’s Dilemma environment, chosen for its capacity to isolate the essential features of strategic interaction while maintaining analytical clarity. This canonical social dilemma captures the conflict between individual incentives and collective welfare, providing a minimal yet expressive context in which to study how information shapes cooperation among artificial agents. The game’s payoff structure follows the standard form: mutual cooperation [C,C] yields 3 points to each player; unilateral cooperation when paired with defection [C,D] or [D,C] yields 5 points to the defector and 0 to the cooperator, an event we define as exploitation of the latter; mutual defection [D,D] yields 1 point to each. 

This environment serves as a strong proof of concept for algorithmic governance for several reasons. First, the Prisoner's Dilemma
most strongly isolates the tension between individual rationality and collective welfare within the class of two-players social dilemmas. This trade-off is a structural property shared by many coordination and resource-allocation problems in multi-agent AI and human governance, making the Prisoner's Dilemma a canonical model for studying how agents balance exploitation and cooperation under uncertainty. Second, its simplicity yields interpretability. Because optimal and suboptimal strategies are well-known analytically characterized, we can precisely trace how informational changes alter decision policies and equilibria. This interpretability is essential for attributing behavioral effects to governance mechanisms rather than to model idiosyncrasies. Furthermore, the PD’s structure is portable: its strategic logic generalizes to a wide range of agentic contexts, from consensus formation and market participation to task allocation. Finally, framing interventions at the informational level rather than through reward shaping or structural redesign provides a limiting-case demonstration of algorithmic governance. If cooperative gains can be achieved solely by modulating what agents know or observe, this suggests a powerful and low-cost governance lever that scales naturally to larger systems without altering their underlying interaction networks.

The experimental setup comprises 20 agents situated within a network, which is initially structured using an Erdos-Renyi model with a 0.25 probability of forming links and keeping fixed within each round, but we evaluate the method over different instances of such networks. Each round of the experiment on each network consists of 20 time steps, during which agents engage in the prisoner's dilemma with adjacent coplayers in their neighborhood sequentially. Due to the relatively small network size, the behavior of LLM agents tends to stabilize within 20 time steps for the majority of the experiments. To ensure robust evaluation, we conduct this simulation across 50 different network topologies, each generated independently using the Erdos-Renyi model.
We utilized LLaMa3-70b LLM, accessed via LangChain and Groq platforms. The model's temperature was set to 0.8. The neural networks employed for the actor and critic components each contain one hidden layer with 256 neurons. The learning rates were configured at 0.001 for the actor and 0.005 for the critic, with a discount factor of 0.99. Both training and evaluation were performed on the High-Performance Cluster (HPC) at Northeastern University.

\subsection{Results from Micro-level Validation of LLM Agent}

To ensure that the experimental results in our project are both reasonable and meaningful, we conducted several micro-level validation tests of the LLM agents' behavior in order to best address our research questions. Based on our findings, we crafted prompts incorporating different types of information for subsequent experiments.

\begin{figure}[t]
    \centering
    \includegraphics[width=0.9\linewidth]{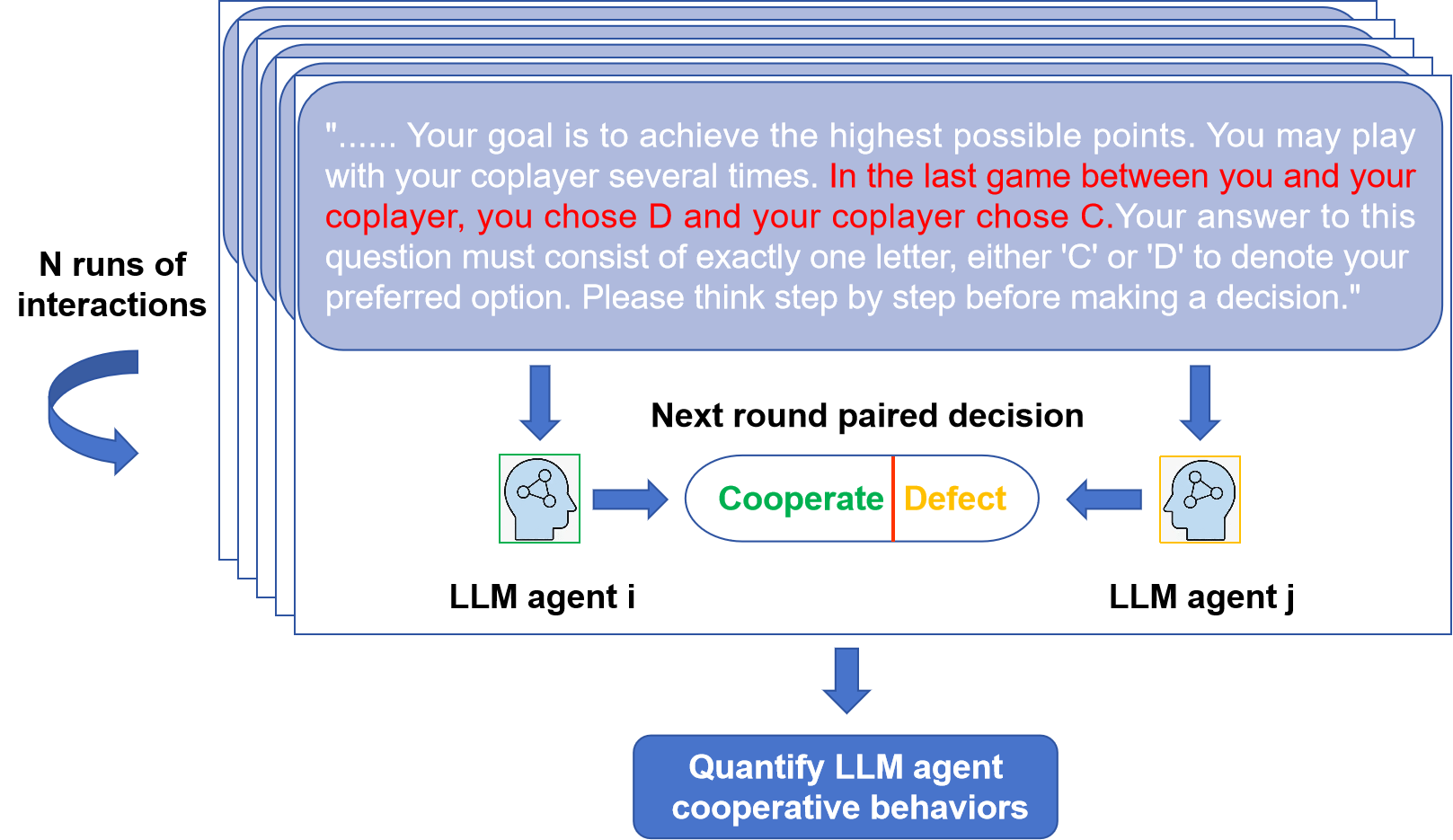}
    \caption{The diagram illustrates how LLM agents interact with each other. The example prompt shown here includes only the information about the last action. For other experimental settings that involve different types of information, additional details will be incorporated into the highlighted (red) sections of the prompts. The LLM agents interact over several rounds, and after collecting their responses, we can compute the cooperation rate under various scenarios with different information conditions. This cooperation rate serves as a key metric for quantifying both the agents’ cooperative behaviors and the overall system performance.}
    \label{fig:prompt_example}
\end{figure}

First, we designed prompts that describe the prisoner’s dilemma and established payoff maximization objectives for the LLM agents to guide their decisions. While it remains uncertain whether LLMs fully understand different games, we avoid explicitly mentioning \enquote{prisoner’s dilemma} to test and fully leverage the agents' autoformalization capabilities. Thus, we presented the payoff matrix sans game labels. Furthermore, we instruct LLM agents to maximize their rewards in the aggregate, noting that it \textit{may} interact with the same co-player multiple times. This is intended to encourage LLM agents to balance short-term gains against and long-term payoffs, much like how humans make strategic decisions based partly on the likelihood of future interactions. Finally, we incorporate \textit{chain-of-thought (CoT)} prompting \cite{wei2022chain} to encourage more strategic reasoning. By default, LLM agents always have access to the most recent action pairs from their interactions with co-players. Figure \ref{fig:prompt_example} shows part of the prompts used when LLM agents have access to the latest action pairs.

After establishing the baseline section of the prompt, we aimed to explore how different types of information influence the LLMs' behavior and to assess whether these influences are both reasonable and consistent. We focused on the three distinct types of information described in Section 3.4: \textbf{1) Last action history (LA)}, \textbf{2) Cooperation ratio of agent and coplayer (AR)}, and \textbf{3) Cooperation ratio of both agent and agent's neighbors (NR)}. These different types of information will be disclosed to the agents depending on the RL manager's decision, occupying the portion in red in the text of Figure \ref{fig:prompt_example}. The placeholder in the prompts like \enquote{\{your\_action\}} or \enquote{\{neighbor\_ratio\}} will be replaced with the real corresponding values in the simulation.

Our initial experiments focused on the Last Action treatment, designed to model the LLMs’ strategic decision-making process in the absence of higher-level contextual information. The results offered a clear picture: LLaMa3-70B consistently reciprocates mutual cooperation, defects after being exploited, and continues to exploit when doing so proves successful. This pattern aligns closely with the well-known “win-stay, lose-shift” (WSLS) or Pavlovian strategy. In Axelrod’s seminal iterated Prisoner’s Dilemma tournaments, the benchmark strategy was Tit-for-Tat (TFT), which simply mirrors the opponent’s last move \cite{axelrod1981evolution}. Subsequent theoretical and experimental work, however, demonstrated that WSLS can outperform TFT in noisy or adaptive environments \cite{Kraines1989Pavlov, Nowak1993Strategy, Wedekind1996WSLS}. In our setting, the model’s tendency to cooperate only 49\% of the time following mutual defection suggests a nuanced variant of WSLS incorporating elements of risk aversion and norm-following: it is comparatively safe to continue cooperating with another cooperator because of the implicit convention it represents, whereas cooperating after mutual defection exposes the model to renewed exploitation. Taken together, these findings suggest that LLMs demonstrate a sophisticated understanding the Prisoner's Dilemma under autoformalization, to the extent that they seem to be able to follow an highly effective and empirically validated strategy.

\begin{figure*}[!htbp]
     \centering
     \includegraphics[width=0.6\linewidth]{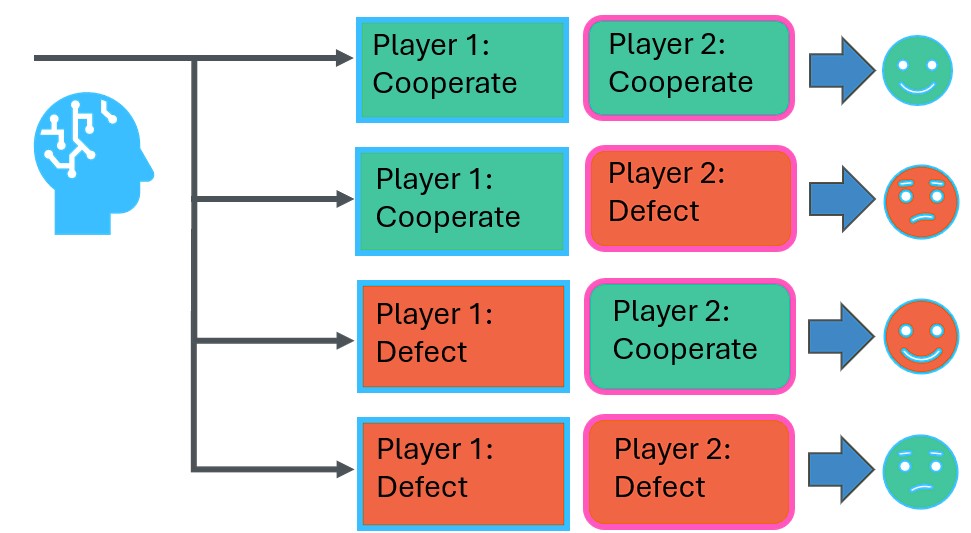}
     \caption{Visual illustration of the Win–Stay, Lose–Shift (WSLS) strategy. Player 1 (light blue) adjusts behavior based on the outcome of the previous round against Player 2 (yellow). When both players cooperate, the outcome is registered as a win, and Player 1 continues to cooperate. When Player 1 is exploited, the strategy registers a loss, prompting a shift to defection. Conversely, after successfully exploiting Player 2, the strategy records a win, and Player 1 maintains defection. Finally, when both players defect, the outcome is registered as a loss, and Player 1 switches to cooperation.}
     \label{WSLS}
\end{figure*}

For our successive analyses, we looked into how LLMs responded when further information on top of LA was disseminated via prompt. Our micro-level validation procedures in these settings demonstrates that Llama3-70b has responds inconsistently to numerical input. To overcome this, we transitioned to qualitative categories for cooperation: We used \textit{Rarely} (to indicate an historical cooperation rate below 33\%), \textit{Sometimes} (for 33\% to 66\%), and \textit{Often} (above 66\%), although this can be extended to higher levels of granularity. \\

Further insight can be derived from studying the tendency of LLM agents to cooperate when Last Action (LA) and Network Ratio (NR) information is available (see SI Section 4 for more details). Consistent with our theory of norm-aware decision making, LLMs never cooperate when cooperation is described as a rare event for both the player and the coplayer. Conversely, following mutual cooperation, LLM agents select cooperation 87\% and 99\% of the time when the coplayer is described as cooperating Sometimes and Often, respectively. Interestingly, when the LLM itself is described as a player who rarely cooperates, it adheres exactly to the win-stay, lose-shift (WSLS) pattern regardless of the coplayer’s cooperation level in the aftermath of exploitation. Finally, after mutual defection, LLM agents slightly increase their probability of switching to cooperation when playing against a coplayer described as Sometimes cooperating. Conversely, when the coplayer is described as cooperating Often, cooperation after mutual defection happens 68\% of the times.

These results are qualitatively similar to those observed when the agent is described as a player who cooperates Sometimes, with only a few key differences. First, except when the coplayer is characterized as cooperating Rarely, the LLM always cooperates following mutual cooperation. Second, when the agent is described as having exploited the coplayer in the previous round and the latter is described as cooperating Often, the LLM chooses to cooperate in roughly one case out of eleven. Lastly, we observe stricter observation of WSLS after mutual defection when the coplayer is described as cooperating Sometimes or Often, once again suggesting that adherence to this strategy is mediated by prevailing norms.

When the agent is described as someone who cooperates Often, we find one particularly salient piece of evidence supporting our claims. If the agent is described as having exploited its coplayer in the previous round and the coplayer is described as cooperating Often, the LLM course-corrects in the following round by choosing cooperation. Although this behavior departs from WSLS prescriptions, it indicates that respect for established norms of reciprocity may take precedence over pure strategic consistency. This interpretation aligns with existing evidence that LLMs act as contextual decision makers, balancing the structural incentives of the game with the social framing of the interaction. Notably, the absence of any difference in the propensity to cooperate after mutual cooperation, whether the agent is described as cooperating Sometimes or Often, suggests diminishing returns to increasingly cooperative framings, with LLMs becoming less responsive once a certain cooperation threshold is reached.

Beyond illustrating the rational consistency of LLMs, these findings highlight a key property: LLMs respond not only to the type of information they receive, but also to its content. It is precisely this nuanced integration of strategic reasoning and contextual sensitivity that makes information modulation an especially effective approach to governing multi-LLM systems.

\subsection{System Performance and Cooperation Rate}

In the previous section, we designed prompts to integrate game structure and information access. Building on this, LLM agents are then constructed, and the RL manager learns to modulate information access for each agent, aiming to optimize the social welfare of all agents. It is important to note that, in many cases, the RL manager can only modulate or expose information from more distant interactions, while the history of recent interactions between an agent and its co-player is expected to be retained in most applications. Thus, we used the \textit{no information} action only for the first period within each round, when LLM agents have no prior interactions with each other. We established several baselines for comparison, each using specific information within the same round. 


\begin{figure*}[t]
    \centering
    \includegraphics[width=0.95\linewidth]{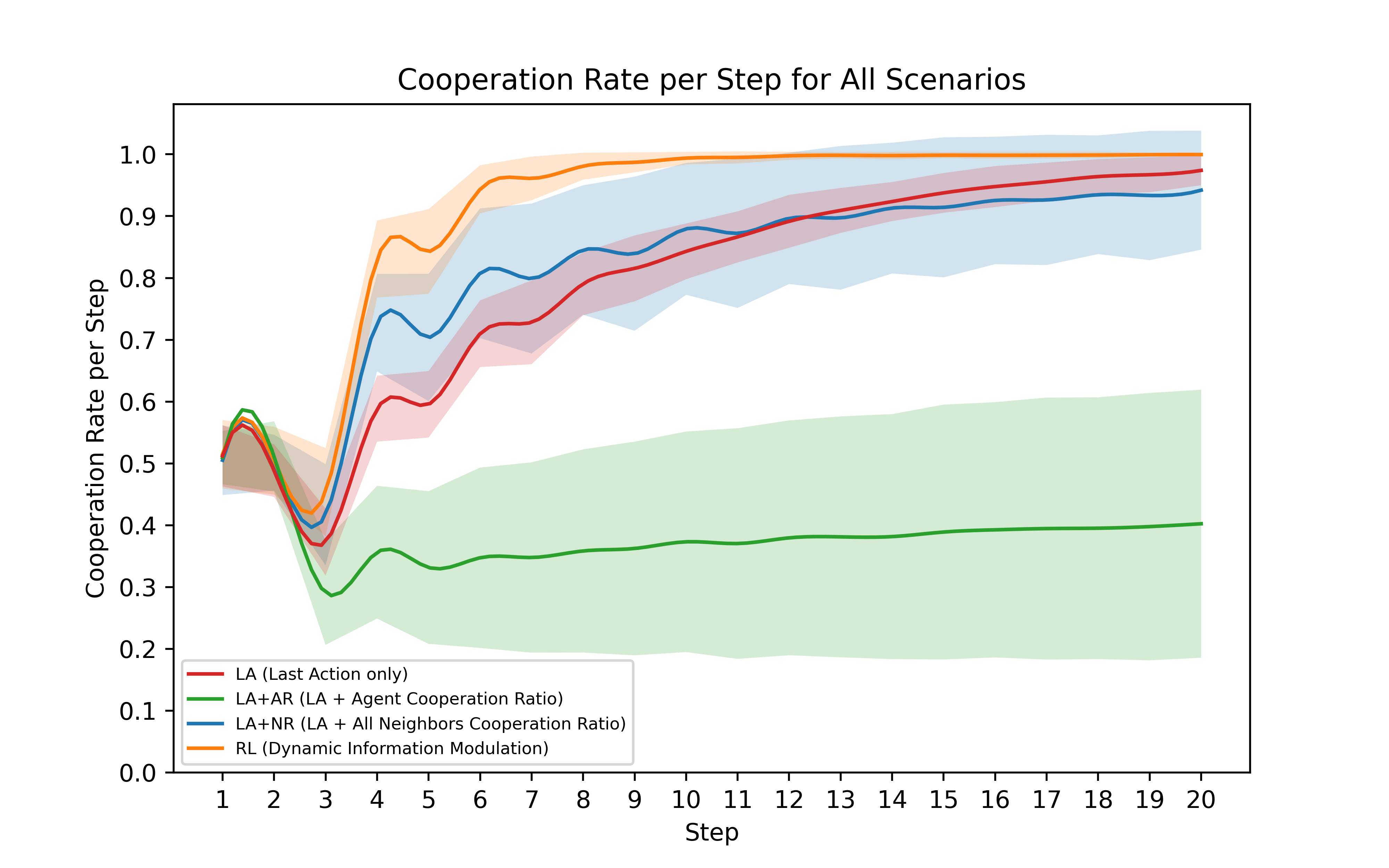}
    \caption{Comparison of cooperation rates over time between RL and baseline methods. The baseline methods utilize specific information during the game, including: \enquote{LA} (last action pairs of both LLM agents), \enquote{LA+NR} (last action pairs of both LLM agents and the overall cooperation ratio of an LLM agent and also of its neighbor agents), and \enquote{LA+AR} (last action pairs and overall cooperation ratio of both LLM agents). The results represent averages from 50 runs, with the shaded areas indicating standard deviation. The results of social welfare in the system over time follow similar trends.}
    \label{cooperation_rate_per_step}
\end{figure*}

In Figure \ref{cooperation_rate_per_step}, we present the normalized social welfare and the system's cooperation rate, averaged across all rounds, with each round involving a different random network instance. Additionally, we report both the time-averaged welfare and cooperation rate across rounds, as well as the outcomes from the final step of the game. We can make the following observations from Figure \ref{cooperation_rate_per_step}: 

First, we observe some expected yet noteworthy results from the static baseline interventions. 
When each LLM agent is only reminded of its most recent interaction with other agents (LA), we see a moderate level of cooperation, consistent with research suggesting that strategies based on recent interaction history can foster cooperative behavior. Extending this by consistently providing each LLM agent with the average cooperation rate of its neighbors further increases the cooperation rate, aligning with findings that cooperation is better promoted within tightly connected clusters of agents \cite{gianetto2015network}. However, providing the overall cooperation rates of individual neighbors yields the lowest social welfare and cooperation levels. This aligns with previous findings that extending memory of past behaviors can diminish cooperation, as it makes agents less forgiving and more prone to defect if a co-player has a history of lower cooperation.

\begin{figure*}[t]
    \centering
    \includegraphics[width=\linewidth]{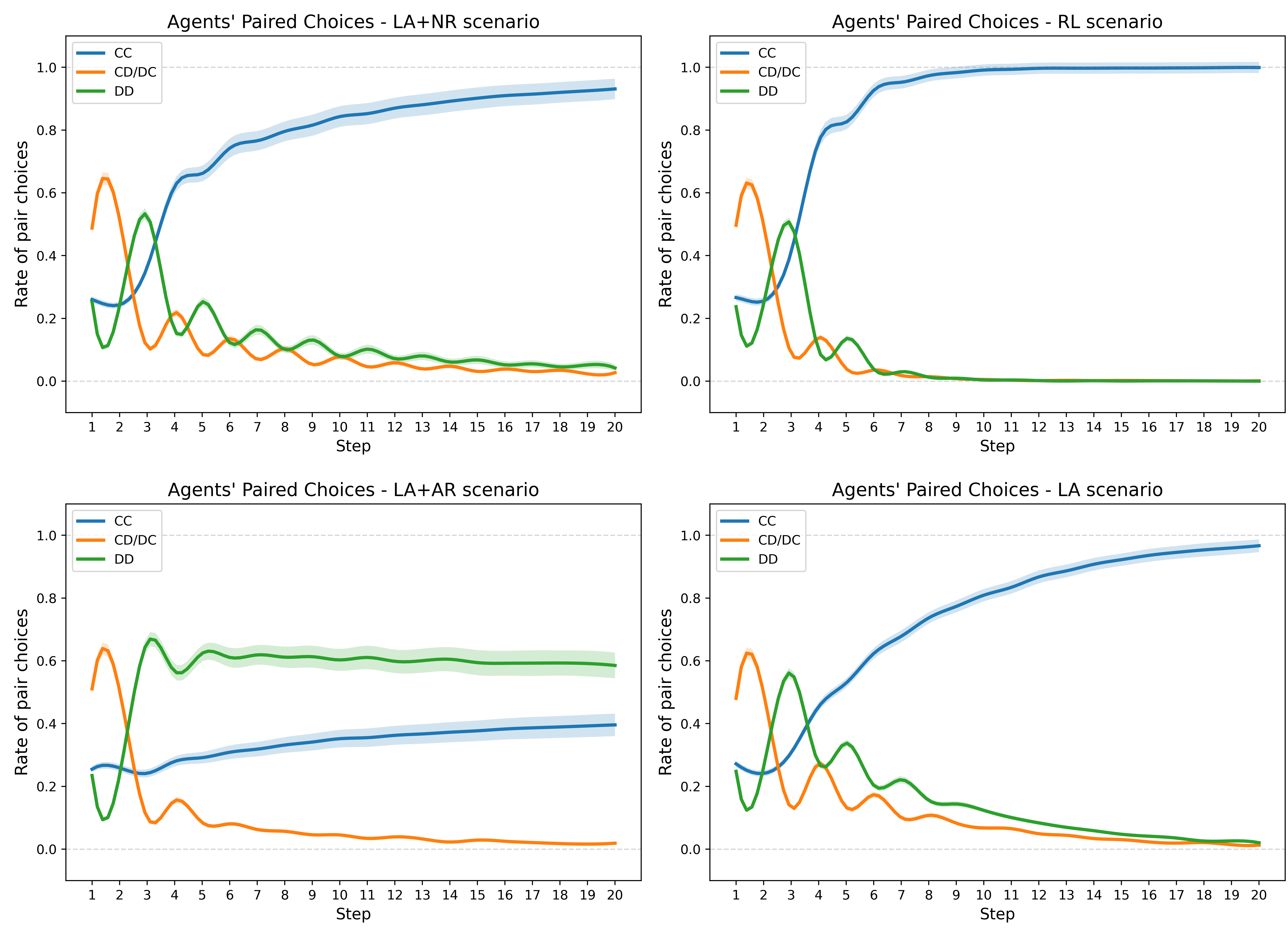}
    \caption{This analysis tracks the evolution of the LLM agent's behavior over time using different methodologies. The Y-axis displays the percentage of different action pairs resulting from interactions. We categorize 'CD' and 'DC' pairs together because they are symmetric and represent equivalent behaviors. The displayed results are averages derived from 50 runs.}
    \label{merged_pair_actions}
\end{figure*}

While these consistent signaling schemes already show considerable variation in performance and cooperation outcomes, our results demonstrate that the RL manager, which dynamically selects among these options, outperforms all baselines across all four metrics. Although the primary objective of the RL manager is to maximize social welfare, its interventions also significantly enhance the system’s cooperation rate, aligning with expectations for the PD game. A 100\% cooperation rate remains unrealistic in real-world applications; however, our findings suggest that RL can effectively learn and adaptively deliver targeted information, boosting both social welfare and cooperation rates in complex systems.

\begin{figure}[h!]
    \centering
    \includegraphics[width=0.95\linewidth]{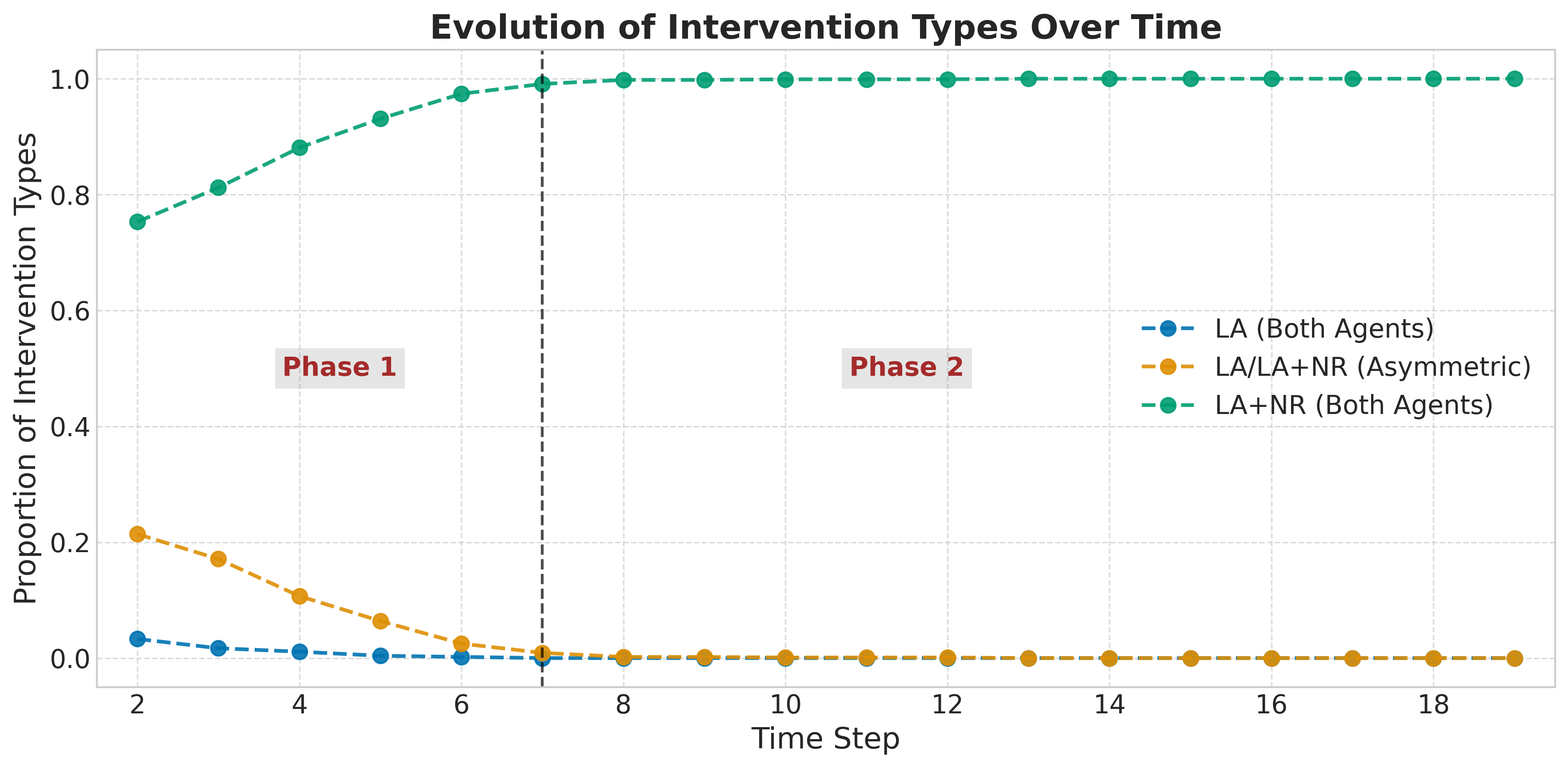}
    \caption{The evolution of intervention types over time in the RL manager’s learned behaviors. The Agent Cooperation Ratio (AR) information is never selected during testing and is thus omitted. \textit{LA (Both Agents)} and \textit{LA+NR (Both Agents)} represent cases where the RL manager assigns the same intervention to both LLM agents within an interaction link, while \textit{LA/LA+NR (Asymmetric)} indicates cases where different information types are assigned to the two LLM agents. The process is divided into two phases to highlight the distinct patterns in the RL manager’s decision-making behavior over time.}

    \label{fig:behavior_PPA}
\end{figure}

\subsection{RL Manager Behavior Analysis}

After demonstrating that the performance of the RL Manager using DRL methods surpasses that of baseline methods, we further analyzed its learned behavior to understand how and why it achieves superior results. 

First, we aim to examine how the behavior of LLM agents evolves over time across different methods. To do this, we analyze the ratios of different action pairs among LLM agents' interaction—[C, C], [C, D], [D, C], and [D, D]—as they change over time. The results are presented in the Figure \ref{merged_pair_actions}. Firstly, the ratio of [C, C] actions for the RL method increases rapidly, reaching 100\% by the 10th timestep. Similar trends are observed for the LA and LA+NR methods, although they increase at a slower pace and do not converge to 100\%. For the AR method, we observe an opposite trend: the ratio of [D, D] actions increases sharply while others decrease. This supports our previous claims that the AR method tends to steer the system toward defection. For the RL, LA, and LA+NR methods, a common trend is observed where the ratio of [D, D] initially increases, reaching a peak in the early steps before subsequently decreasing. This pattern begins as LLM agents, initially devoid of memory, randomly choose between C (cooperate) and D (defect) under the 'no prior information' prompts. This randomness often results in pairs of [C, D] or [D, C], which can escalate into a predominance of [D, D] outcomes. This occurs as agents strive to avoid being exploited by their co-players while also seeking to gain an advantage themselves.


Second, we observe that the LA and LA+NR baselines achieve relatively good performance from Figure \ref{cooperation_rate_per_step}. However, the RL manager still outperforms them, indicating that it does not simply learn to always choose one of these strategies. Instead, it appears to have learned a more sophisticated policy. We can gather supporting evidence from Figure \ref{fig:behavior_PPA}, which illustrates the evolution of different intervention strategies over time based on the RL manager's learned behavior. First, we observe that for the majority of the time, the RL manager chooses to reveal the neighborhood cooperation rate to the LLM agents. This aligns with expectations, as this information alone is sufficient to achieve strong performance in many different situations.

From Figure \ref{fig:behavior_PPA}, we can easily separate the process into 2 phases: phase 1 (steps 2-6) and phase 2 (steps 7 and later).
In phase 1, the RL manager learns to rely more on the LA (last action) information to some extent. As discussed in section 4.2, when the last action pair is [D, D] and the LLM agents are situated in a neighborhood prone to defection, revealing the neighborhood cooperation rate may reinforce defection, potentially leading to nearly 100\% defection. In contrast, providing only the last action pair information in such scenarios can lead LLM agents to cooperate approximately 50\% of the time, making it more effective than revealing the LA+NR (neighborhood cooperation rate). This demonstrates that the RL manager learns to selectively use last action information in specific situations, which is a key reason it outperforms the LA+NR baseline. In phase 2, the RL manager consistently selects the neighborhood cooperation rate, indicating convergence toward a stable and effective policy over time.

Additionally, the figure clearly shows that the RL manager occasionally assigns different types of information to LLM agents within the same interaction link, demonstrating its ability to make appropriate heterogeneous decisions when necessary. It reveals an intriguing pattern of strategic information provision that emerges naturally from the manager's reinforcement learning process. Starting from step 2, LA+NR information already dominates at approximately 75\% of interventions, but the system maintains a significant proportion (around 20\%) of mixed interventions until step 6. This asymmetric information strategy appears to be a sophisticated approach to managing defection dynamics in the early stages. By selectively revealing heterogeneous information types to agents within the same interaction link, the RL manager is capable of helping break potential deadlocks of mutual defection and triggering cooperative behaviors. The complete transition to network-based information by step 7 (marked by the vertical dashed line in the figure) suggests that once cooperation patterns are sufficiently established in the network, RL manager learns to reveal uniform information provision to agents, which can maintain mutual cooperation. 

This analysis explains the RL manager's superior performance compared to other deterministic scenarios. The strategic allocation of information demonstrates the RL manager's ability to make nuanced decisions about when and to whom different information types should be revealed. In the following sections, we examine how specific agent characteristics (degree centrality and cooperation history) influence the RL manager's information allocation decisions. We investigate whether agents with varying network positions and cooperative behavior in previous games receive systematically different information levels, using a T-test analysis to determine the statistical significance of these differences. Our analysis reveals that the RL manager employs a sophisticated strategy where both network topology and agents' behavioral histories significantly shape intervention targeting, showing how the system leverages these dual characteristics to optimize cooperation across the entire network.

\subsubsection{Heterogeneous Decision with Different Degree Centrality}

\begin{figure}[h!]
    \centering
    \includegraphics[width=0.6\linewidth]{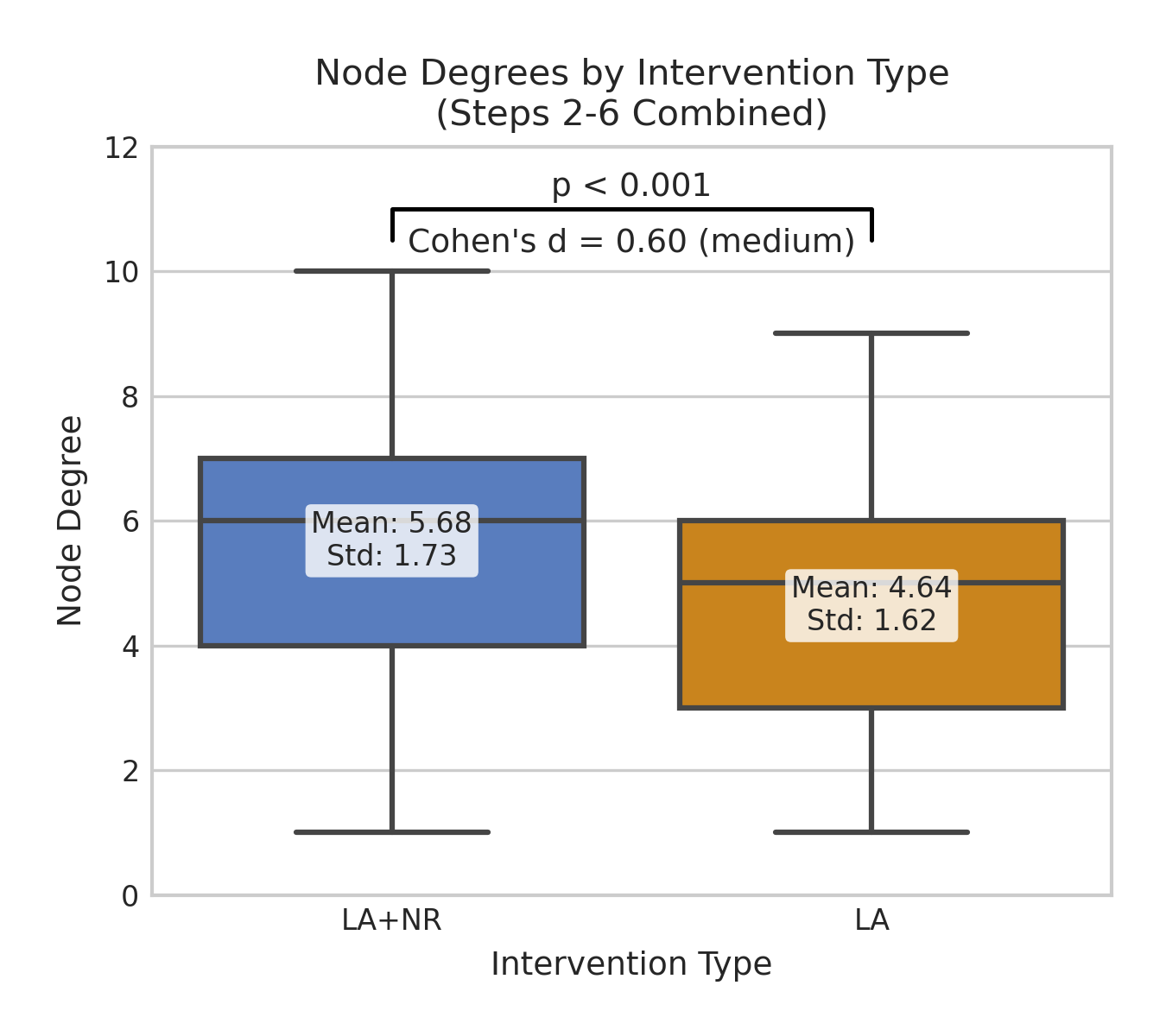}
    \caption{This figure shows the relationship between node degree (number of connections) and intervention types during the critical early steps (2-6). Agents receiving LA+NR information (neighborhood ratio) have significantly higher connectivity (mean degree: 5.68) compared to those receiving LA information only (mean degree: 4.64). This indicates that the RL manager strategically allocates network-level information to more central nodes in the network. Steps 7 and beyond are excluded since we only have the dominant information, LA+NR, for that phase.}
    \label{fig:steps2to6_node_degree_comparison}
\end{figure}

Based on the statistical analysis of node degrees by intervention type in Figure \ref{fig:steps2to6_node_degree_comparison}, an interesting pattern emerges. Nodes with higher degrees (average 5.68) are significantly more likely to receive neighborhood ratio information, while nodes with lower degrees (average 4.64) tend to receive last action information. This difference is statistically significant ($p < 0.001$) with a medium effect size (Cohen's d = 0.60), suggesting this is not a random occurrence but rather a deliberate strategy that emerged from the reinforcement learning process \cite{cohen2013statistical}. We did this test for steps 2 to 6 only, since after that, all agents are receiving the dominant strategy, neighborhood ratio, and the comparison of intervention types is meaningless.

This finding reveals a significant correlation: the RL manager, despite having no knowledge of network structure, exhibits a pattern where more central agents receive richer, network-level information while peripheral agents receive simpler information (i.e., LA only). This emergent allocation pattern proves functionally effective because high-degree nodes can amplify intervention impacts throughout the network. When a well-connected agent adopts cooperative behavior based on neighborhood information, their influence reaches many other agents. Conversely, providing basic information to less connected agents may suffice for their decision-making, given their limited network influence. While the manager lacks explicit awareness of network topology, this differentiated information allocation emerges as an efficient approach that maximizes system-wide cooperation. The t-test results confirm this correlation is statistically significant, suggesting that the RL manager has converged on an information distribution pattern that, though unintentionally, aligns with network-theoretic principles of influence propagation.

\subsubsection{Heterogeneous Decision with Different Cooperation History}

\begin{figure}[!htbp]
    \centering
    \includegraphics[width=0.6\linewidth]{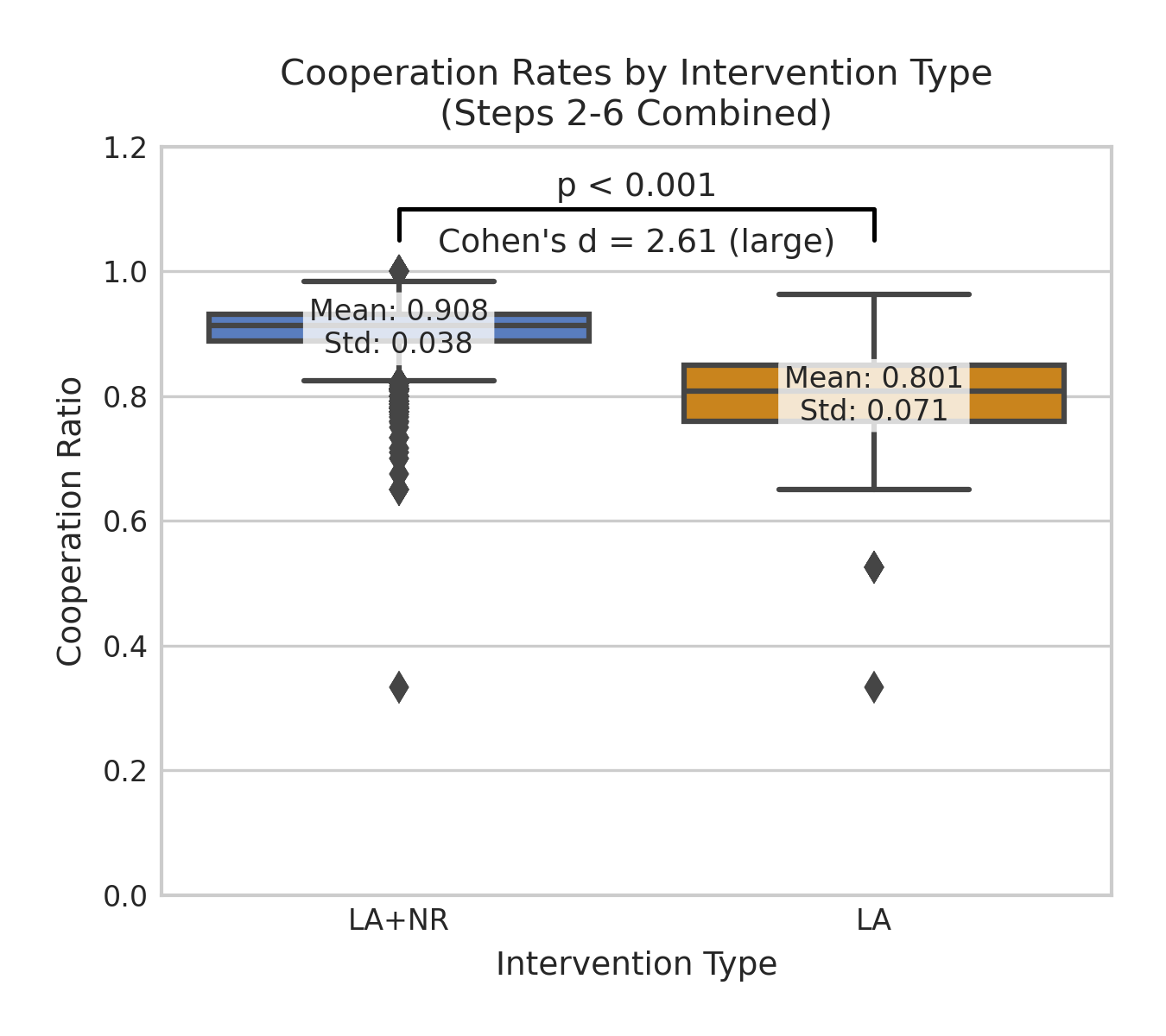}
    \caption{This figure shows how the RL manager's intervention decisions correlate with agents' prior cooperation history. Agents receiving LA+NR information show substantially higher pre-intervention cooperation rates (mean: 0.908) than those receiving LA information (mean: 0.801). This reveals that the RL manager selectively targets last action information toward agents with lower cooperation tendencies, while providing network-level information to already cooperative agents.}
    \label{fig:steps2to6_coop_ratio_comparison}
\end{figure}

Analysis of the cooperation rate in Figure \ref{fig:steps2to6_coop_ratio_comparison} reveals a significant difference between intervention types. Agents receiving LA+NR information demonstrate significantly higher cooperation rates before intervention (mean: 0.908) compared to those receiving LA information (mean: 0.801). This difference is both statistically significant ($p < 0.001$) and represents a remarkably large effect size (Cohen's d = 2.61). Notably, the LA+NR group also shows less variability in cooperation, suggesting more consistent cooperative behavior. This shows that the RL manager strategically assigns different information types based on agents' cooperation history. Rather than randomly distributing information, the system preferentially provides the simpler LA information to agents with lower cooperation tendencies, while agents who demonstrated cooperative behavior receive the optimal LA+NR information. It seems that for less cooperative agents, it is more influential to provide LA information only, to foster their cooperative behavior.

These findings, combined with our node degree analysis and intervention evolution pattern over time, reveal the complicated strategy employed by the RL manager. During the critical early steps (2-6), the RL manager makes discriminating choices about information provision based on both network position and cooperation history. It targets LA information toward less cooperative agents and those with fewer connections, potentially as a remedial strategy to improve their behavior without reinforcing negative patterns that might emerge from network-level information. Meanwhile, it leverages the cooperative tendencies of well-connected nodes by providing them with LA+NR information, allowing them to maintain and spread their cooperative behavior. This nuanced approach demonstrates how the RL manager learns to tailor interventions to individual agent characteristics, explaining its superior performance compared to one-size-fits-all strategies.

\section{Conclusion}
In this paper, we propose a framework comprising multiple Strategic LLM agents positioned in a random network, interacting with their neighbors, and an RL manager that dynamically provides information to LLM agents to foster pro-social behavior and maximize social welfare. Each LLM agent receives prompts, including descriptions of pairwise strategic games, objectives, and additional information from the RL manager, to make decisions such as cooperation or defection. The information set and prompts are refined through micro-level validation to ensure LLM agents' behaviors are consistent and reasonable. The manager, trained via reinforcement learning, observes relevant information from both LLM agents in each interaction and determines the optimal level of information to provide, aiming to maximize social welfare. The evaluation results demonstrate that the manager with RL outperforms other baseline methods in various aspects. Furthermore, the analysis of the learned behavior of RL manager generates meaningful insights.

This paper takes an initial step toward exploring a governance framework using deep reinforcement learning (DRL) for multi-LLM systems, an increasingly important class of multi-agent systems in real-world applications \cite{estornell2024multi}. The proposed concept of information modulation enhances the realism of the governance framework by influencing agent behavior through the modulation of perceptions, rather than by directly enforcing actions. Indeed, governance in our model hinges on intervening on the \textit {information} layer, rather than on the \textit{interaction} layer of the network, an approach that comes with a distinct set of advantages. By constraining the manager to modulating information only, we accommodate for scenarios and applications in which a manager cannot directly intervene on the connections between agents, which may be exogenous and given or arising from bottom-up processes that a manager could not realistically orchestrate. As a direct consequence, we also derive a governance scheme whose efficacy is not predicated on the underlying interaction network. Finally, limiting intervention to the information allows for scalable solutions, as modifying information flows is often less costly than altering the network structure itself. Finally, this work opens up a promising direction for simulating real-world multi-agent systems that involve human agents. 

However, this work presents a few limitations. First, the limited sample size of the evaluation (number of rounds) may introduce fluctuations in the observed trends, primarily due to constraints in computational resources. However, the confidence intervals in our results indicate that increasing the number of rounds is unlikely to alter the key findings significantly. A potential solution to this issue is using small-scale LLMs fine-tuned by large models, which can replicate certain behaviors of the larger models while being more computationally efficient. Additionally, future work could explore different network structures beyond random networks, test the framework across other strategic games, and incorporate more granular information tiers for RL manager's intervention.



\newpage
\listoffigures
\listoftables

\end{document}